\documentclass[10pt,twocolumn,letterpaper]{article}

\usepackage{cvpr}              
\usepackage[accsupp]{axessibility}  

\usepackage{graphicx}
\usepackage{amsmath}
\usepackage{amssymb}
\usepackage{booktabs}
\usepackage{bbding}
\usepackage{pifont}
\usepackage{color}
\usepackage{tabularray}
\definecolor{Iron}{rgb}{0.87,0.878,0.89}
\newcommand{\rightsign}{\ding{51}}
\newcommand{\wrongsign}{\ding{55}}

\definecolor{Color1}{RGB}{255, 153, 153}
\definecolor{Color2}{RGB}{255, 204, 153}
\definecolor{Color3}{RGB}{254, 248, 173}

\definecolor{cvprblue}{rgb}{0.21,0.49,0.74}
\usepackage[pagebackref,breaklinks,colorlinks,allcolors=cvprblue]{hyperref}

\usepackage[capitalize]{cleveref}
\crefname{section}{Sec.}{Secs.}
\Crefname{section}{Section}{Sections}
\Crefname{table}{Table}{Tables}
\crefname{table}{Tab.}{Tabs.}

\def\paperID{10232} 
\def\confName{CVPR}
\def\confYear{2025}

\begin{document}

\title{Sparse Point Cloud Patches Rendering via Splitting 2D Gaussians }

\author{Changfeng Ma$^1$, Ran Bi$^1$, Jie Guo$^1$, Chongjun Wang$^1$, Yanwen Guo$^{12*}$\\
$^{1}$Nanjing University, Nanjing, China $^{2}$School of Software, North University of China\\
{\tt\small \{changfengma, 211250233\}@smail.nju.edu.cn}\\
{\tt\small \{guojie,chjwang,ywguo\}@nju.edu.cn}
}

\twocolumn[{%
\renewcommand\twocolumn[1][]{#1}%
\maketitle
\vspace{-2em}
\centering
\includegraphics[width=0.9\linewidth]{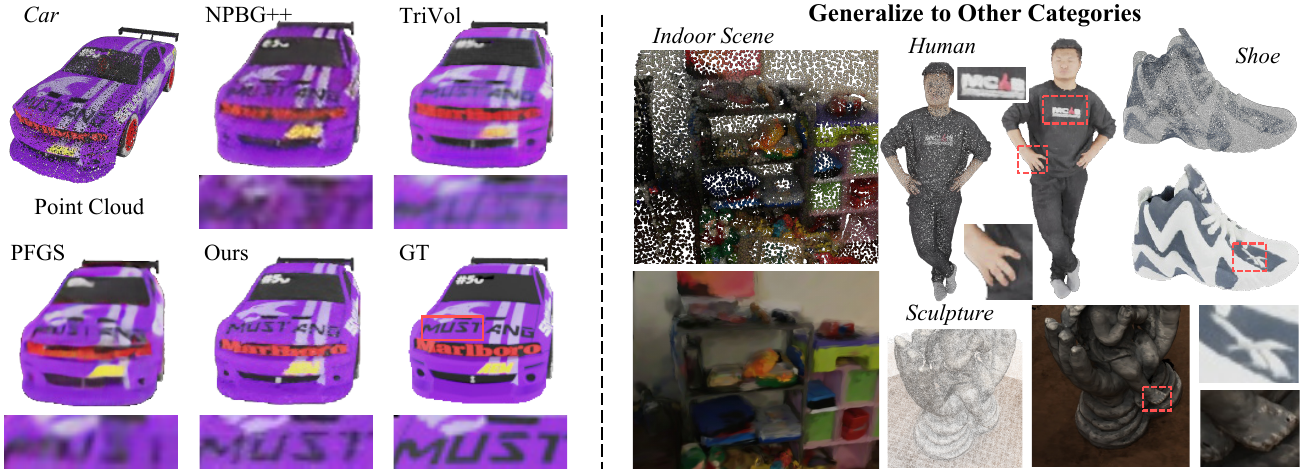}
\captionof{figure}{Our method excels in rendering intricate, photo-realistic images from point clouds of different categories. In this instance, our model is trained on the Car category, utilizing 20K points as input. 
\vspace{2em}
}
\label{fig:teaser}
}]

\renewcommand{\thefootnote}{} 
\footnotetext{$^*$Corresponding author.}

\begin{abstract}
Current learning-based methods predict NeRF or 3D Gaussians from point clouds to achieve photo-realistic rendering but still depend on categorical priors, dense point clouds, or additional refinements.
Hence, we introduce a novel point cloud rendering method by predicting 2D Gaussians from point clouds.
Our method incorporates two identical modules with an entire-patch architecture enabling the network to be generalized to multiple datasets.
The module normalizes and initializes the Gaussians utilizing the point cloud information including normals, colors and distances.
Then, splitting decoders are employed to refine the initial Gaussians by duplicating them and predicting more accurate results, making our methodology effectively accommodate sparse point clouds as well.
Once trained, our approach exhibits direct generalization to point clouds across different categories. 
The predicted Gaussians are employed directly for rendering without additional refinement on the rendered images, retaining the benefits of 2D Gaussians.
We conduct extensive experiments on various datasets, and the results demonstrate the superiority and generalization of our method, which achieves SOTA performance.
The code is available at \href{https://github.com/murcherful/GauPCRender}{https://github.com/murcherful/GauPCRender}.
\end{abstract}

\section{Introduction}
\label{sec:intro}

Point clouds can be easily acquired using various types of 3D scanners, making them a recent research hotspot.
Photo-realistic rendering of point clouds is significant for many applications such as visualization, virtual reality and automatic driving.  
However, the sparsity and discrete nature of point clouds continue to pose challenges for achieving photo-realistic rendering.

Traditional graphics rendering methods \cite{GR1, GR2} merely project points as planes or spheres onto images. Consequently, their outcomes lack photo-realism and are riddled with numerous holes.
Previous deep learning-based works \cite{NPBG,NPCR,NPBG++} necessitate point clouds and corresponding images from several views as input, which are difficult to acquire in practice.
These methods also require training or fine-tuning for each scene or object.
Recent works \cite{TriVol,Point2Pix} have focused on predicting NeRF \cite{NeRF} from given point clouds to render images from arbitrary views.
PFGS \cite{PFGS} predicts the 3D Gaussians \cite{3DGS} with features for point clouds. After rendering, PFGS refines the rendered images based on these features.
Although these methods can render point clouds without training or fine-tuning on each data, they still have limitations, such as the need for dense point clouds, poor generalization, slow rendering speeds, and blurry results.

\begin{table}[t]
\centering
\footnotesize
\caption{The comparison of different methods across several dimensions. Here, ``S. S/O.'', ``S. C.'' and ``M. C.'' represent ``single scene or object'', ``single category'' and ``multiple categories'', respectively. $\mathcal{P}$ and $\mathcal{I}$ indicate point clouds and images, separately. $\dagger$: Traditional graphics rendering approach is unable to render photo-realistic images. $\ddagger$: PFGS requires further refinement on images rendered from 3D gaussians. }
\vspace{-2mm}
\label{tab:relate_compare}
\begin{tblr}{
  cells = {c},
  colsep = 3pt,
  column{1} = {leftsep=1pt,rightsep=2pt},
  column{2,4} = {leftsep=1pt,rightsep=1pt},
  column{5} = {rightsep=1pt},
  column{6} = {leftsep=1pt},
  vline{2} = {-}{0.07em},
  hline{3} = {-}{dashed},
  hline{6} = {-}{dashed},
  hline{1,10} = {-}{0.12em},
  hline{2,9} = {-}{0.07em},
}
Method    & \begin{tabular}[c]{@{}c@{}}Generalization \\ability\end{tabular}& Input & Output & \begin{tabular}[c]{@{}c@{}}Pre- \\train\end{tabular} & \begin{tabular}[c]{@{}c@{}}Fine- \\tune\end{tabular} & \begin{tabular}[c]{@{}c@{}}Point \\number\end{tabular}            \\
GR$^\dagger$        & Any           & $\mathcal{P}$    & $\mathcal{I}$      & \wrongsign       & \wrongsign        & Any                     \\
NPBG \cite{NPBG}      & S. S/O.       & $\mathcal{P}$+$\mathcal{I}$  & $\mathcal{I}$      & \rightsign      & \rightsign       & 100K                    \\
NPCR \cite{NPCR}     & S. S/O.       & $\mathcal{P}$+$\mathcal{I}$  & $\mathcal{I}$      & \wrongsign       & \rightsign       & 100K                    \\
NPBG++ \cite{NPBG++}   & S. S/O.       & $\mathcal{P}$+$\mathcal{I}$  & $\mathcal{I}$      & \rightsign      & \wrongsign  & 100K                    \\
TriVol \cite{TriVol}   & S. C.         & $\mathcal{P}$    & NeRF\cite{NeRF}   & \rightsign      & \wrongsign        & 100K                    \\
Point2Pix \cite{Point2Pix} & S. C.         & $\mathcal{P}$    & NeRF\cite{NeRF}   & \rightsign      & \wrongsign        & 100K                    \\
PFGS  \cite{PFGS}      & S. C.         & $\mathcal{P}$    & 3DGS\cite{3DGS}$^\ddagger$   & \rightsign      & \wrongsign        & 80K                     \\
Ours      & M. C.         & $\mathcal{P}$    & 2DGS\cite{2DGS}   & \rightsign      & \wrongsign        & 2K-100K 
\end{tblr}
\vspace{-2mm}
\end{table}

2D Gaussian Splatting \cite{2DGS} has recently been proposed, which includes explicit normals of Gaussians.
These normals facilitate easy initialization, learning, and prediction by the network from point clouds.
Motivated by this, we introduce a novel method to predict the 2D Gaussians of the given point cloud and rasterize the predicted 2D Gaussians into images for photo-realistic point cloud rendering.

Due to taking entire point clouds as input, previous methods are hard to generalize to other categories or adapt to different point distributions.
Processing only point cloud patches with the network can enhance its generalization capability and reduce dependence on category priors \cite{poco, hu2019randla}.
However, rendering Gaussians predicted from patch points results in incomplete images that are unsupervisable.
Therefore, we employ an entire-patch architecture that includes two identical 2D Gaussian prediction modules to handle entire and patch point clouds, separately.
We use the rendering results of the entire point cloud as a ``background'' to obtain a complete image for proper supervision.
The module first normalizes and initializes the 2D Gaussians based on the given point clouds, encompassing the estimated normals, scales, and colors, which are crucial for our network to converge and predict more accurate results.
Subsequently, a point cloud encoder and multiple splitting decoders are employed to extract local features and predict the parameters of the duplicated Gaussians.
Equipped with the splitting decoder, our network can handle sparse point clouds and produce high-quality details with dense Gaussians.
Thanks to the proposed architecture, our method also exhibits strong generalization across scenes and objects.
Without further refinement on rendered images, our method retains the advantages of 2D Gaussian Splatting in rendering, such as rapid rendering speed.

We conduct extensive comparative and ablation experiments, and the results demonstrate the superiority of our method, which can render point clouds into high-quality detailed images with excellent generalization.
We will make our dataset and codes public in the future. 
Our main contributions are as follows:
\begin{itemize}
    \item We introduce a method that directly predicts 2D Gaussians from point clouds to render photo-realistic images without any refinement, while retaining the advantages of 2D Gaussian Splatting.

    \item Equipped with an entire-patch architecture, our method can handle point cloud patches, leading to significant generalization across scenes and objects.

    \item We introduce the splitting decoder, which splits Gaussians into denser outcomes, enabling our method to render detailed images from even sparse point clouds.

\end{itemize}

\section{Related Work}
\subsection{Point Cloud Rendering} \label{sec:related_work_1}

\textbf{Traditional Point Cloud Rendering.} 
Point cloud rendering has consistently been a significant research topic \cite{GraphsicRenderSurvey}.
Conventional point cloud rendering techniques \cite{pytorch3d,open3d,pyvista,meshlab} directly rasterize points onto 2D screens for rendering, yet they grapple with the issue of holes.
Subsequent studies \cite{GR1,GR2,GR3,GR4} have proposed splatting points with elliptic discs, ellipsoids, or surfels for point cloud rendering to mitigate the holes observed during the rendering process.
Although these graphic rendering methods can render any point cloud, the resulting images often contain holes and exhibit blurry details, lacking photo-realism.

\noindent\textbf{Point-based novel view synthesis.}
For photo-realistic point cloud rendering, several recently proposed methods utilize corresponding images of point clouds as additional information. 
NPBG \cite{NPBG} signs descriptors to point clouds and rasterizes into multi-resolution raw images with features and employs a U-Net to render the final results.
Dai et al. \cite{NPCR} aggregate points with features onto multiple planes based on a given view and point cloud. Following multi-plane-based voxelization, they employ a 3D U-Net to predict the rendered image.
ADOP \cite{ADOP} is introduced for rendering HD images from point clouds, which employs several global parameters, a neural renderer and a differentiable physically-based tonemapper for optimization.
These methods necessitate training for each point cloud with images from several views to render images of novel views, which is a time-consuming process.
To reduce training time, NPBG++ \cite{NPBG++} is trained on a collection of point clouds using a network that extracts features from corresponding images. This network aligns the features of input images to point clouds without the need for fine-tuning.
However, this method still requires images as input, which can be challenging to obtain in practice.

\begin{figure*}[t]
    \centering
    \includegraphics[width=\linewidth]{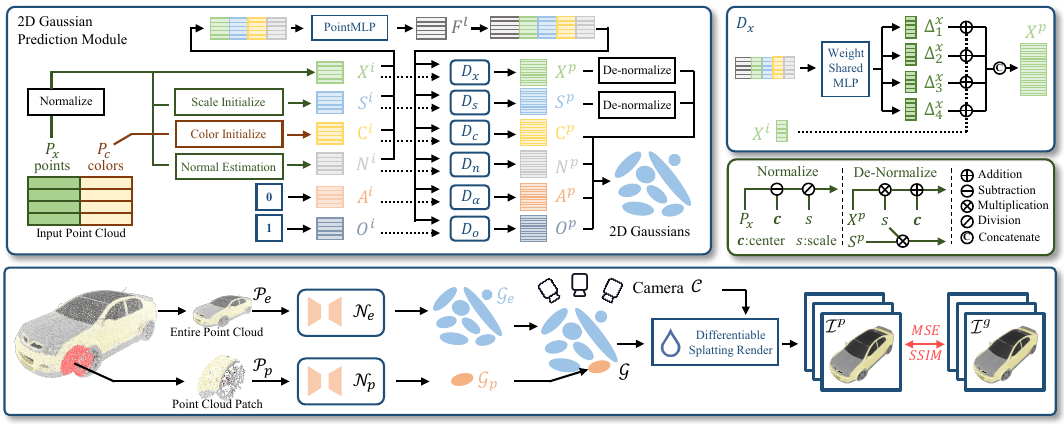}
    \vspace{-1em}
    \caption{Overview of our proposed method. Our method predicts 2D Gaussians for point cloud rendering, employing an entire-patch architecture (bottom) and the 2D Gaussian Prediction Module (top left) with splitting decoders (top right).}
    \vspace{-1em}
    \label{fig:method}
\end{figure*}

\noindent\textbf{Learning-based Point Cloud Rendering.}
Several works \cite{TriVol,Point2Pix,PFGS} are proposed for rendering point clouds without the need for training or fine-tuning on each point cloud and corresponding images.
TriVol \cite{TriVol} utilizes triple-plane-based grouping to query point features for predicting a NeRF from a given point cloud. 
Hu et al. \cite{Point2Pix} propose multi-scale radiance fields to predict NeRFs from point clouds. They also introduce a sampling strategy to accelerate fine-tuning and rendering speed.
PFGS \cite{PFGS}, a two-stage approach, for the first time employs feature splatting to predict 3D Gaussians from point clouds and uses a recurrent decoder to refine the rendered images.
These methods are capable of rendering point clouds without fine-tuning or the need for input images, yet they still depend on categorical priors, dense points, or further refinements.
Table \ref{tab:relate_compare} summarizes the distinctions between our method and the  previous methods across several dimensions, highlighting the innovation and superiority of our proposed approach.

\subsection{Gaussian Splatting}
3D Gaussian Splatting \cite{3DGS} is proposed with remarkable performance in novel view synthesis and further explores the rendering capability of the point clouds. 
3DGS utilizes the differentiable splatting rendering to optimize 3D Gaussian primitives initialized from point clouds.
Scenes are explicitly represented with 3D Gaussians that can be rendered with remarkable speed.
Meshes can also be reconstructed from 3D Gaussians\cite{sugar}. 
However, the normals of 3D Gaussians are so vague that the resulting geometries lack precision.
To generate more accurate geometries, Hua et al. \cite{2DGS} propose 2DGS that employs 2D Gaussian disks containing explicit normals as representation primitives. 
With precise normals, meshes can accurately reconstruct the geometry of objects.
Given that the normals of 2D Gaussians can be readily determined, 2D Gaussians can also be initialized using normals.
Luckily, the normals of points can be effortlessly estimated \cite{Hoppe1992SurfaceRF} facilitating the initialization and forecasting of 2D Gaussians.
Motivated by this, we opt to predict 2D Gaussians in place of 3D Gaussians from the provided point clouds.

\section{Method}
Given a point cloud $\mathcal{P}$ and a camera $\mathcal{C}$, point cloud rendering aims to render a photo-realistic image $\mathcal{I}$.
Here, point cloud $\mathcal{P}$ contains of coordinates of 3D points $\mathbf{P_x} \in \mathbb{R}_{N\times 3} $ and colors of points $\mathbf{P_c} \in [0, 1]_{N\times 3}$, where $N$ indicates the point number of $\mathcal{P}$.
Camera $\mathcal{C}$ consists of intrinsic matrix $\mathbf{K} \in \mathbb{R}_{3 \times 3}$ and extrinsic matrix $\mathbf{P} \in \mathbb{R}_{4 \times 4}$.

For photo-realistic point cloud rendering, we propose a network $\mathcal{N}$ to predict 2D Gaussians $\mathcal{G}=\mathcal{N}(\mathcal{P})$ given point cloud $\mathcal{P}$, and employ differentiable splatting rendering \cite{2DGS} $f_{R}$ to synthesize images $\mathcal{I} = f_{R}(\mathcal{G}, \mathcal{C})$. 
The parameters of a 2D Gaussian $\mathbf{g} = \{\mathbf{x}, \mathbf{s}, o, \mathbf{c}, \mathbf{n}, \alpha \} \in \mathcal{G}$  include: (1) position $\mathbf{x} \in \mathbb{R}_3$, (2) scale $\mathbf{s} \in \mathbb{R}_2 $, (3) opacity $ o \in [0, 1]$, (4) spherical harmonic of color $\mathbf{c} \in \mathbb{R}_d$, (5) normal $\mathbf{n} \in \mathbb{R}_3$ and (6) rotation angle along normal $\alpha \in [0, 2\pi]$, where $d$ represents the parameter number of spherical harmonic for color.  
As shown in Figure \ref{fig:method}, our network contains two identical 2D Gaussion prediction modules (\ref{sec:GPM}) with an entire-patch architecture (\ref{sec:epa}).

\begin{figure}[t]
    \centering
    \includegraphics[width=\linewidth]{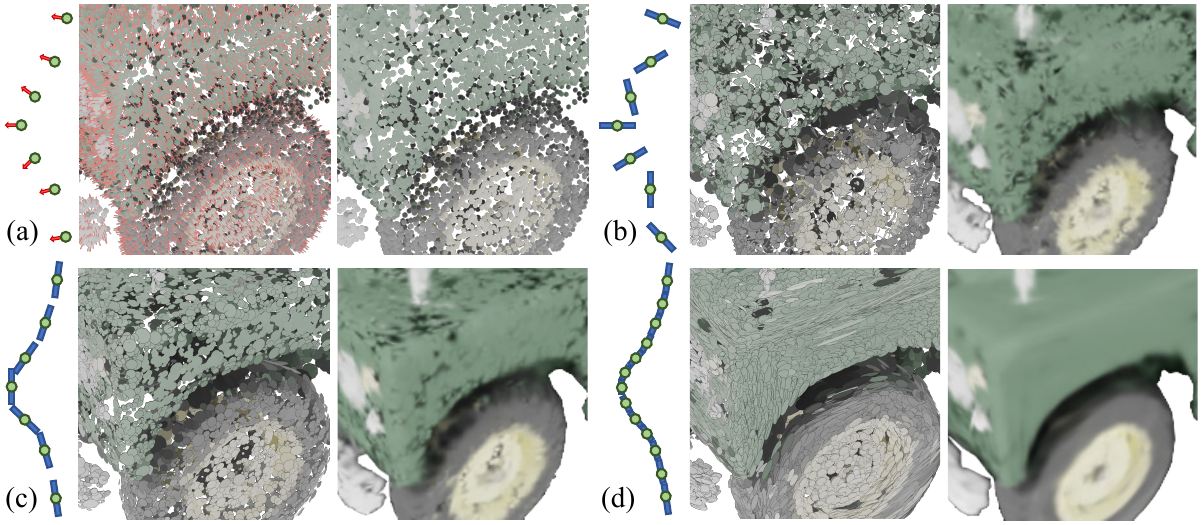}
    \vspace{-1.5em}
    \caption{The illustration of our initialization approach, where the left side of each picture is a 2D schematic diagram and the right side is the rendered images of Gaussians. (a) The estimated normals of the point cloud. (b) Randomly Initialization. (c) Our Initialization. (d) Predicted 2D Gaussians. }
    \vspace{-1.5em}
    \label{fig:init}
\end{figure}

\subsection{2D Gaussian Prediction Module}\label{sec:GPM}
\subsubsection{Initialization of 2D Gaussians}
We first introduce 2D Gaussian prediction module to predict the 2D Gaussians $\mathcal{G}$ given a point cloud $\mathcal{P}$ as input. 
The positions of the point cloud $\mathcal{P}$ are normalized to the range $[-1, 1]$ based on the geometric center $\mathbf{c} \in \mathbb{R}_3$ and scale $s\in R$ of the input point cloud to facilitate easier feature extraction and  prediction of results, where the scale is defined as the maximum distance between any point and $\mathbf{c}$.
Our module then initializes one 2D Gaussian $\mathbf{g}$ for each normalized point $\mathbf{p} = \{\mathbf{p_x}, \mathbf{p_c}\} \in \mathcal{P}$, where $\mathbf{p_x} \in \mathcal{R}_3$ and $\mathbf{p_c} \in \mathcal{R}_3$ indicate the position and color of the point respectively. 
The position $\mathbf{x}$ of the Gaussian is initialized with $\mathbf{p_x}$.
The spherical harmonic $\mathbf{c}$ and scale $\mathbf{s}$ are initialized as the same as 2DGS \cite{2DGS} according to $\mathbf{p_c}$ and the minimum distance between points, separately.
The opacity $o$ is set to $1$ to ensure that all Gaussians are visible.
As we mentioned before, the geometry of given point cloud $\mathcal{P}$ provides important clues for the rotation of Gaussians.
Gaussian Splatting works \cite{3DGS, 2DGS} utilize quaternions to represent the rotation of Gaussians and only randomly initialize them as Figure \ref{fig:init}(b) shows, which are difficult for networks to predict intuitively.
Fortunately, the normals of given point cloud can be easily estimated. 
Therefore, we represent the rotation of the Gaussian in terms of the normal $\mathbf{n}$ and the rotation angle $\mathbf{\alpha}$ around that normal, and initialize the normal of the Gaussian using the estimated normal of the point cloud.
Figure \ref{fig:init} (a) and (c) illustrate this initialization process. 
Specifically, we estimate the normals for $\mathcal{P}$ \cite{Hoppe1992SurfaceRF, open3d}, where the normal $\hat{\mathbf{n}}$ for each point $\mathbf{p}$ is utilized to initialize the normal $\mathbf{n}$ of Gaussian. And the rotation angle $\alpha$ of Gaussian is set to $0$.
The estimated normal may not be precise, but it provides crucial information and ensures that each Gaussian is visible, thereby guaranteeing that the network can converge correctly.
After initializing for all the points, the initialized Gaussians $\mathcal{G}^i$ composing of the initialized positions, normals, scales, spherical harmonics, opacities and rotation angles ($\mathbf{X}^i,\mathbf{N}^i \in \mathbb{R}_{N\times3}, \mathbf{S}^i\in \mathbb{R}_{N\times2}, \mathbf{C}^i \in \mathbb{R}_{N\times3}, O^i, A^i \in \mathbb{R}_N$) is passed to our network as input and predicting basis.

\vspace{-0.5em}
\subsubsection{Splitting Decoder}
\vspace{-0.5em}
As shown in Figure \ref{fig:method}, local features are required before splitting and predicting 2D Gaussians.
To extract features from the initialized 2D Gaussians, we employ PointMLP\cite{PointMLP} as the encoder $E$, which is a simple residual MLP framework for point cloud learning with great performance on many tasks.
The local feature $\mathbf{F}_l$ is extracted by:
\begin{equation*}
    \mathbf{F}_l = E (\mathbf{X}^i, \mathbf{C}^i, \mathbf{N}^i, \mathbf{S}^i).
\end{equation*}
Subsequently, $\mathbf{F}_l$ is augmented with $\mathbf{X}^i, \mathbf{C}^i, \mathbf{N}^i, \mathbf{S}^i$ as the input features for the splitting decoders.
Our splitting decoders $D$ also utilize $\mathcal{G}^i$ as the basis for prediction and forecast 2D Gaussians $\mathcal{G}^p$ for rendering purposes.
There are six splitting decoders $D_x, D_s, D_c, D_n, D_\alpha, D_o$ responsible for predicting the six parameters $X^p, S^p, C^p, N^p, A^p, O^p$ of 2D Gaussians.
Each decoder contains a weight-shared MLP \cite{2017PointNet} as shown in Figure \ref{fig:method}.
The splitting and predicting process for the 6 parameters is basically the same, except for slight differences in shapes.
Take the process of predicting the position $X^p$ as an example. 
The decoder $D_{x}$ predicts $K$ shifts $\mathbf{\Delta}^x$ based on the position $\mathbf{X}^i$:
\begin{equation*}
    \mathbf{\Delta}^x_1, \mathbf{\Delta}^x_2, ..., \mathbf{\Delta}^x_K = D_{x}(\mathbf{F}_l, \mathbf{X}^i, \mathbf{C}^i, \mathbf{N}^i, \mathbf{S}^i),
\end{equation*}
where $K$ indicates the number of splits.
The $K$ shifts are added to the same $\mathbf{X}^i$ that is ``split" the initial position $\mathbf{X}^i$ to $K$ new positions.
The predicted position $\mathbf{X}^p \in \mathbb{R}_{K\cdot N\times3}$ is the combination of these $K$ new positions:
\begin{equation*}
    \mathbf{X}^p = \bigcup_{j=1}^{K}(\mathbf{\Delta}^x_j + \mathbf{X}^i).
\end{equation*}
After applying all the splitting decoders, we can obtain the parameters of $K\cdot N$ Gaussians $\mathcal{G}^p$. 
With the splitting decoders, our method can manage sparse point clouds by augmenting the number of Gaussians, serving as a form of point cloud upsampling.
A greater number of Gaussians can also capture more complex features, enhancing the details of rendered images.
The predicted 2D Gaussians are based on normalized points, hence they need to be de-normalized to achieve accurate rendering results.
The de-normalization is applied to the predicted position $X^p$ and scale $S^p$.

\subsection{Entire-Patch Architecture}\label{sec:epa}
However, the 2D Gaussian prediction module can only be trained on complete point clouds, which leads to poor generalization across different categories with varying point distributions, as we only have ground truth images of entire objects for supervision.
Numerous methods for other point cloud tasks, including surface reconstruction\cite{poco} and semantic segmentation\cite{hu2019randla}, utilize point cloud patches as processing units to enhance the generalization capabilities of their networks.
However, the corresponding ground truth images of point cloud patches are unavailable in practice, which makes training on patches impossible.
To facilitate the application of our method to point cloud patches, we utilize an entire-patch architecture, as illustrated at the bottom of Figure \ref{fig:method}.
Our approach encompasses two 2D Gaussian prediction modules, $\mathcal{N}_e$ and $\mathcal{N}_p$, which operate on the entire point cloud and the point cloud patch, respectively.
Given an entire point cloud $\mathcal{P}_e$ for rendering, our method randomly selects a point as the center point of a point cloud patch. 
The point cloud patch $\mathcal{P}_p$ is obtained by gathering $N_p$ nearest points around the center point.
The 2D Gaussians of point cloud $\mathcal{P}_e$ and patch $\mathcal{P}_p$ are predicted by two prediction modules, where $\mathcal{G}_e = \mathcal{N}_e(\mathcal{P}_e)$ and $\mathcal{G}_p = \mathcal{N}_p(\mathcal{P}_p)$. 
The non-patch part $\mathcal{G}_e'$ of $\mathcal{G}_e$ are the 2D Gaussians corresponding to the non-patch points $\mathcal{P}_e' = \mathcal{P}_e - \mathcal{P}_p$.
The final 2D Gaussians $\mathcal{G} = \mathcal{G}_e' \cup \mathcal{G}_p$ are used to render images for supervision during training.
Here, $\mathcal{G}_e'$ serves as the ``background" ensuring that the rendered image is complete for calculating the loss function  and $\mathcal{N}_p$ can be properly trained.
In practice, we initially train $\mathcal{N}_e$ separately and then freeze $\mathcal{N}_e$ while training $\mathcal{N}_p$ for our method.
Both $\mathcal{N}_e$ and $\mathcal{N}_p$ can be utilized for inference.
The inference of $\mathcal{N}_e$ directly takes $\mathcal{P}_e$ as input and outputs the entire 2D Gaussians for rendering.
For $\mathcal{N}_p$, our method divides the point cloud into several patches.
To ensure that all points of the input point cloud are covered, we repeatedly perform the process of randomly selecting a central point and extracting surrounding points as the patch point cloud, continuing until all points are included in at least one patch.
The 2D Gaussian of the entire point cloud is the combination of the predictions from all patches.



\subsection{Training Details}
Given $N_c$ cameras $\mathcal{C}_1, \ldots , \mathcal{C}_{N_c}$, the rendered images $\mathcal{I}^p_1, \ldots, \mathcal{I}^p_{N_c}$ is obtained by $f_R(\mathcal{G}, \mathcal{C}_1), \ldots, f_R(\mathcal{G}, \mathcal{C}_{N_c})$.
Similar to \cite{2DGS}, we employ MSE and SSIM between rendered images and the corresponding ground truth images $\mathcal{I}^g_1, \ldots, \mathcal{I}^g_{N_c}$ as loss functions for supervision:
\begin{equation*}
    \mathcal{L} = \frac{1}{N_c}\sum_{i=1}^{N_c} \big(\beta\mathcal{L}_{MSE}(\mathcal{I}^p_i, \mathcal{I}^g_i) + (1-\beta)\mathcal{L}_{SSIM}(\mathcal{I}^p_i, \mathcal{I}^g_i)\big).
\end{equation*}

We implement our method with PyTorch \cite{pytorch}.
The split number $K$ is set to 4.
The point number $N_p$ of a patch is 2048.
$\beta$ is set to 0.8 for the loss function.
The number of rendered images $N_c$ is 8. 
The resolutions of images are 256x256, 640x512 and 512x512 for objects, scenes and human bodies. 
We employ Adam optimizer whose learning rate is $1.0\times 10^{-4}$. The batch size is set to $8$, and the maximum epoch of training is $480$. Training our method takes about $30$ hours with a RTX 3090 GPU. 


\section{Experiments}

\subsection{Settings}
\textbf{\textit{Dataset}.}
We compare our method with previous works across 5 datasets that include scenes, objects, and human bodies, encompassing the following datasets.
\textbf{ShapeNet} \cite{ShapeNet} is a large-scale dataset that contains CAD models across various categories.
\textbf{GSO} \cite{GSO} is a high-quality dataset comprising 3D scanned household items.
\textbf{ScanNet} \cite{ScanNet} is a dataset consisting of real-scanned indoor scenes.
\textbf{DTU} \cite{DTU} is a multi-view stereo dataset featuring high-quality and high-density scenes.
\textbf{THuman2.0} \cite{THuman} is a dataset comprising high-quality 3D models of human bodies.
We select the chairs and tables from ShapeNet and shoes from GSO for object-level comparison. 
All settings are consistent with those of TriVol \cite{TriVol} and PFGS \cite{PFGS}.

\begin{figure}[t]
    \centering
    \includegraphics[width=\linewidth]{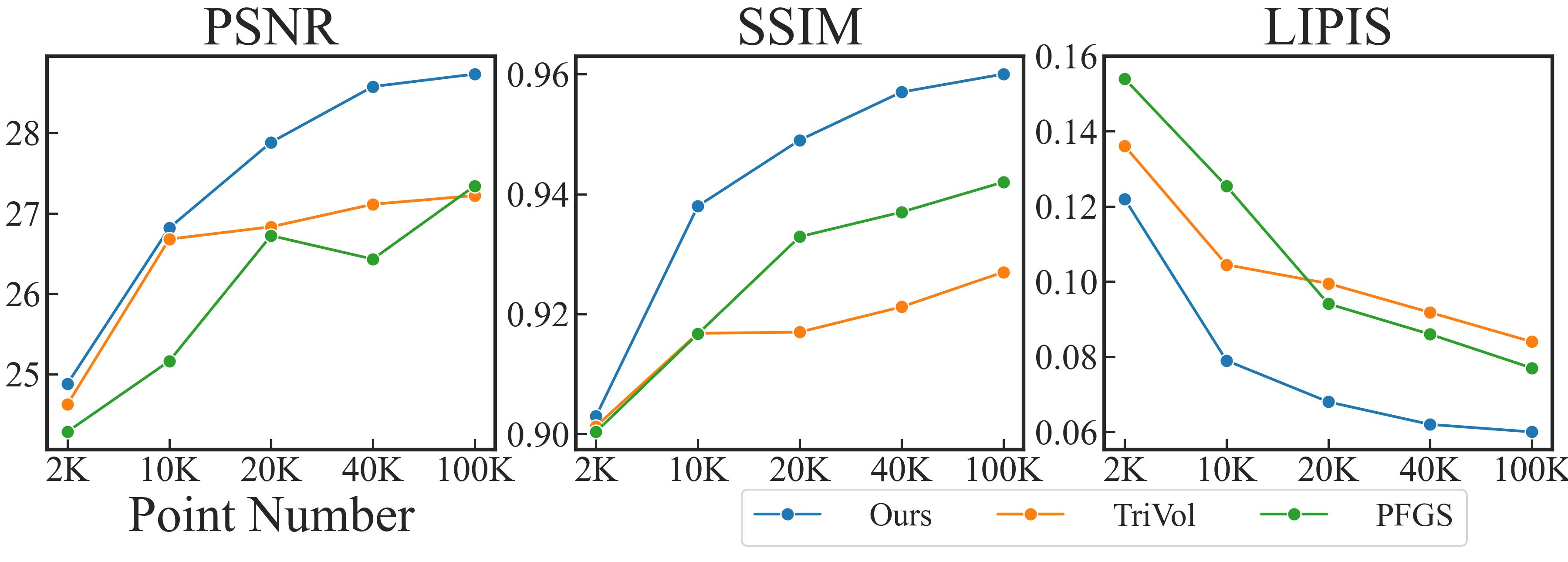}
    \caption{The evaluation of our method, TriVol and PFGS trained with different point numbers on Car category. The legend in the lower right corner indicates different methods. }
    \label{fig:diff_num}
\end{figure}

\noindent\textbf{\textit{Metrics}.} We evaluate the rendering results using three widely accepted metrics: PSNR, SSIM\cite{SSIM}, and LIPIS\cite{Lipis}.

\noindent\textbf{\textit{Baselines}.} We employ traditional graphics rendering (\textbf{GR}), point-based novel view synthesis method \textbf{NPBG++}\cite{NPBG++} and learning-based method \textbf{TriVol}\cite{TriVol} and \textbf{PFGS}\cite{PFGS} for comparison on different datasets.

\begin{figure*}[t]
    \centering
    \includegraphics[width=\linewidth]{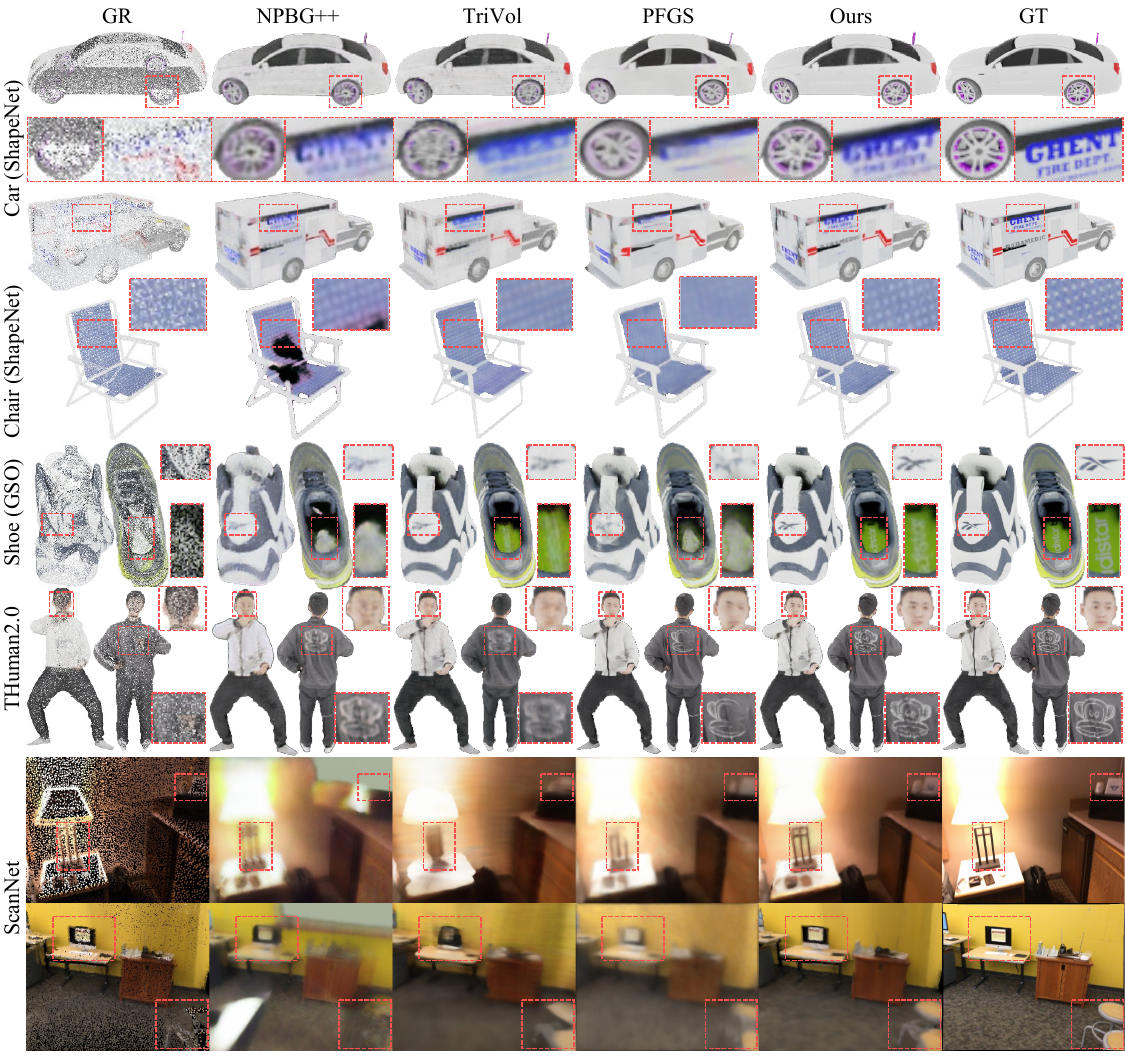}
    \vspace{-1.5em}
    \caption{The rendering results of our method and previous methods on different categories. From top to bottom: scene, car, chair, shoe and human body.}
    \label{fig:compare}
    \vspace{-1.5em}
\end{figure*}




\begin{table*}
\footnotesize
\centering
\caption{The comparison of our method against previous methods on different datasets including scenes, objects and human bodies. 
The top three results are highlighted in red, orange, and yellow, respectively.
$\dagger$:The input point number for THuman2.0 is 80K for all methods.}
\label{tab:main_com}
\begin{tblr}{
  cells = {c},
  cell{1}{1} = {r=2}{},
  cell{1}{2} = {r=2}{},
  cell{1}{3} = {c=3}{},
  cell{1}{6} = {c=3}{},
  cell{1}{9} = {c=3}{},
  cell{1}{12} = {c=3}{},
  cell{1}{15} = {c=3}{},
  cell{7}{1} = {r=3}{},
  cell{10}{1} = {r=3}{},
  vline{2,3,6,9,12,15} = {-}{0.07em},
  hline{1,10} = {-}{0.12em},
  hline{2} = {3-17}{0.07em},
  hline{3,7} = {-}{0.07em},
  hline{4,5} = {-}{dashed},
  colsep = 1.75pt,
  cell{6}{5}={Color1},cell{6}{17}={Color1},cell{8}{9}={Color1},cell{8}{10}={Color1},cell{8}{11}={Color1},cell{9}{3}={Color1},cell{9}{4}={Color1},cell{9}{6}={Color1},cell{9}{7}={Color1},cell{9}{8}={Color1},cell{9}{12}={Color1},cell{9}{13}={Color1},cell{9}{14}={Color1},cell{9}{15}={Color1},cell{9}{16}={Color1},cell{9}{17}={Color1},
  cell{5}{5}={Color2},cell{5}{14}={Color2},cell{6}{3}={Color2},cell{6}{4}={Color2},cell{6}{15}={Color2},cell{6}{16}={Color2},cell{8}{6}={Color2},cell{8}{7}={Color2},cell{8}{8}={Color2},cell{8}{12}={Color2},cell{8}{13}={Color2},cell{9}{9}={Color2},cell{9}{10}={Color2},cell{9}{11}={Color2},
  cell{5}{12}={Color3},cell{7}{6}={Color3},cell{7}{7}={Color3},cell{7}{8}={Color3},cell{7}{9}={Color3},cell{7}{10}={Color3},cell{7}{11}={Color3},cell{7}{13}={Color3},cell{8}{3}={Color3},cell{8}{4}={Color3},cell{8}{14}={Color3},cell{8}{15}={Color3},cell{8}{16}={Color3},cell{8}{17}={Color3},cell{9}{5}={Color3},
}
Method & {Point\\Number} & ScanNet\cite{ScanNet} &       &       & Car (ShapeNet\cite{ShapeNet}) &       &       & Chair (ShapeNet\cite{ShapeNet}) &       &       & Shoe (GSO\cite{GSO}) &       &       & THuman2.0\cite{THuman} &       &       \\
       &                 & PSNR$\uparrow$     & SSIM$\uparrow$  & LPIPS$\downarrow$  & PSNR$\uparrow$           & SSIM$\uparrow$  & LPIPS$\downarrow$ & PSNR$\uparrow$             & SSIM$\uparrow$  & LPIPS$\downarrow$ & PSNR$\uparrow$       & SSIM$\uparrow$  & LPIPS$\downarrow$ & PSNR$\uparrow$      & SSIM$\uparrow$  & LPIPS$\downarrow$ \\
GR                  & 100K$^\dagger$            & 13.62  & 0.528 & 0.779 & 19.24         & 0.814 & 0.182 & 17.78           & 0.779 & 0.201 & 23.14     & 0.829 & 0.153 & 20.26    & 0.905 & 0.337 \\
NPBG++\cite{NPBG++} & 100K            & 16.81  & 0.671 & 0.585 & 25.32         & 0.874 & 0.120 & 25.78           & 0.916 & 0.101 & 29.42     & 0.929 & 0.081 & 26.81    & 0.952 & 0.062 \\
TriVol\cite{TriVol} & 100K            & 18.56  & 0.734 & 0.473 & 27.22         & 0.927 & 0.084 & 28.85           & 0.960 & 0.078 & 31.24     & 0.961 & 0.045 & 25.97    & 0.935 & 0.059 \\
PFGS\cite{PFGS}     & 100K            & 19.86  & 0.758 & 0.452 & 27.34         & 0.942 & 0.077 & 27.52           & 0.956 & 0.078 & 29.53     & 0.957 & 0.058 & 34.74    & 0.983 & 0.009 \\
Ours          & 20K             & 18.57  & 0.724 & 0.600 & 27.88         & 0.949 & 0.068 & 28.89           & 0.962 & 0.077 & 30.91     & 0.968 & 0.051 & 33.70    & 0.979 & 0.014 \\
                    & 40K             & 19.43  & 0.743 & 0.552 & 28.57         & 0.957 & 0.062 & 29.52           & 0.967 & 0.070 & 31.49     & 0.974 & 0.047 & 34.29    & 0.981 & 0.012 \\
                    & 100K            & 20.24  & 0.759 & 0.490 & 28.73         & 0.960 & 0.060 & 29.10           & 0.965 & 0.075 & 32.08     & 0.978 & 0.042 & 35.43    & 0.987 & 0.009 \\
\end{tblr}
\end{table*}

\subsection{Evaluation of Rendering}
The evaluation results of our method and baselines are presented in Table \ref{tab:main_com}.
The point numbers $N$ of input point clouds are 100K for objects and scenes, and 80K for human bodies.
The results demonstrate that our method outperforms nearly all other methods across various metrics and categories.
Our approach outperforms existing methods, showing average improvements of 2.60\%, 1.05\%, and 6.06\% in PSNR, SSIM, and LIPIS across all datasets, respectively.
For qualitative comparison, we represent the rendering results of different methods across several datasets in Figure \ref{fig:compare}. 
Our rendering results not only exhibit superior overall quality, but also display crisp and detailed elements, such as patterns, wheels, and human faces.
Quantitative and qualitative comparisons demonstrate that our method achieves SOTA performance in point cloud rendering.

\subsection{Evaluation on Different Point Number}
As shown in Table \ref{tab:main_com}, we also display the evaluation results of our method trained with different input point numbers, including 20K and 40K.
Despite utilizing fewer points, our method still achieves the best performance on object categories such as chairs, tables, and shoes.
Note that our method only utilizes 20\% of the points as input, indicating that our method also yields good results for sparse point clouds.
We also showcase the evaluation results for TriVol and PFGS when trained with varying point counts (2K, 10K, 20K, 40K, 100K) on the Car category to illustrate their performance with sparse point clouds.
The comparisons are depicted in Figure \ref{fig:diff_num}.
All methods exhibit improved performance as the number of points used during training increases, with our method outperforming all others across all input point numbers.
The comparisons highlight the superiority of our method in rendering sparse point clouds.

\begin{table}
\footnotesize
\centering
\caption{
The generalization capability of our method, traditional graphics rendering, and PFGS on the DTU dataset.
Here, our method is trained on the Car category of ShapeNet using 20K points. 
Two models of PFGS are trained on the DTU dataset with 1M points and on the Car category of ShapeNet with 20K points, respectively.
The first column records the input point numbers for evaluation. 
The top two results are highlighted in red and orange.
}
\label{tab:gen_com}
\begin{tblr}{
  cells = {c},
  cell{1}{1} = {r=2}{},
  cell{1}{2} = {c=3}{},
  cell{1}{5} = {c=3}{},
  cell{7}{1} = {r=2}{},
  cell{7}{2} = {c=3}{},
  cell{7}{5} = {c=3}{},
  vline{2,5} = {1-12}{0.07em},
  hline{1,13} = {-}{0.12em},
  hline{2,8} = {2-7}{0.07em},
  hline{3,7,9} = {-}{0.07em},
  colsep = 4.2pt,
  cell{3}{2} = {Color2},
  cell{4}{2-4} = {Color2},
  cell{5}{2} = {Color1},cell{5}{3} = {Color2},cell{5}{4} = {Color2},
  cell{6}{2-4} = {Color1},
  cell{9}{3,4} = {Color2},
  cell{9,10}{5} = {Color1},
  cell{9-11}{6,7} = {Color1},
  cell{11}{5} = {Color2},
  cell{12}{5-7} = {Color2},
}
{ Point \\Number} & {PFGS \\ DTU 1M points} &          &           & {PFGS \\ Car (ShapeNet) 20K points} &          &           \\
                  & PSNR$\uparrow$                       & SSIM$\uparrow$ & LPIPS$\downarrow$ & PSNR$\uparrow$                         & SSIM$\uparrow$ & LPIPS$\downarrow$ \\
20K               & 10.177                         & 0.313    & 0.647     & 8.065                            & 0.297    & 0.792     \\
40K               & 12.026                         & 0.358    & 0.550     & 8.160                            & 0.302    & 0.765     \\
100K              & 17.978                         & 0.554    & 0.312     & 8.647                            & 0.322    & 0.673     \\
1M                & 23.001                         & 0.770    & 0.112     & 13.676                           & 0.549    & 0.401     \\
{ Point \\Number} & GR                             &          &           & \begin{tabular}[c]{@{}c@{}}Ours \\ Car (ShapeNet) 20K points\end{tabular}  &          &           \\
                  & PSNR$\uparrow$                       & SSIM$\uparrow$ & LPIPS$\downarrow$ & PSNR$\uparrow$                         & SSIM$\uparrow$ & LPIPS$\downarrow$ \\
20K               & 9.105                          & 0.316    & 0.625     & 16.713                           & 0.590    & 0.361     \\
40K               & 9.985                          & 0.328    & 0.641     & 16.623                           & 0.582    & 0.315     \\
100K              & 12.035                         & 0.363    & 0.666     & 16.544                           & 0.573    & 0.263     \\
1M                & 18.003                         & 0.594    & 0.396     & 16.643                           & 0.594    & 0.216   
\end{tblr}
\end{table}

\begin{figure}[t]
    \centering
    \includegraphics[width=\linewidth]{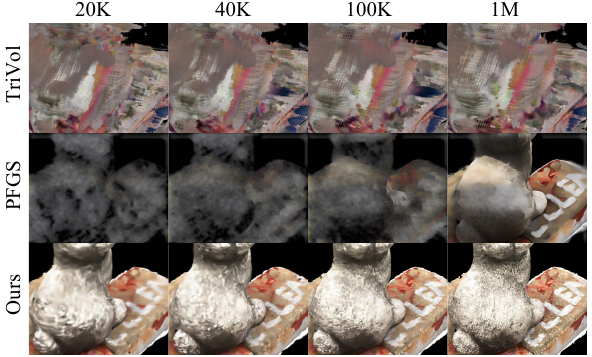}
    \caption{The evaluation results of our method and previous methods on the DTU dataset, utilizing 20K, 40K, 100K and 1M points, with all methods trained on the Car category with 20K points.
    }
    \label{fig:gener_dc_dpn}
\end{figure}

\subsection{Evaluation of Generalization Capability}
To evaluate the generalization capability of our $\mathcal{N}_p$ on different categories with different input point numbers, we directly test our network on DTU dataset, which is trained on the Car category using 20K points . 
The setting here is the same as that in PFGS \cite{PFGS}.
The results are presented in Table \ref{tab:gen_com}, where the methods are assessed using 20K, 40K, 100K and 1M points on the DTU dataset.
We also present the evaluation results of PFGS when trained on DTU with 1M points and on the Car category of ShapeNet with 20K points, in order to compare the generalization capability of PFGS.
Like previous methods, PFGS requires entire point clouds as input for predicting Gaussians, hence these methods struggle to generalize to other datasets or varying point densities.
Being trained with point cloud patches, our method does not depend on global priors and possesses superior generalization capabilities, resulting in enhanced performance on other datasets and with varying point densities.
Figure \ref{fig:gener_dc_dpn} shows the results of methods on DTU dataset with different point numbers, where our rendering results preserve clear details.
Note that previous methods are unable to function on sparse point clouds or on scene categories that are significantly different from object categories.
Both quantitative and qualitative evaluations reveal the exceptional generalization capability of our method.

\begin{figure}[t]
    \centering
    \includegraphics[width=\linewidth]{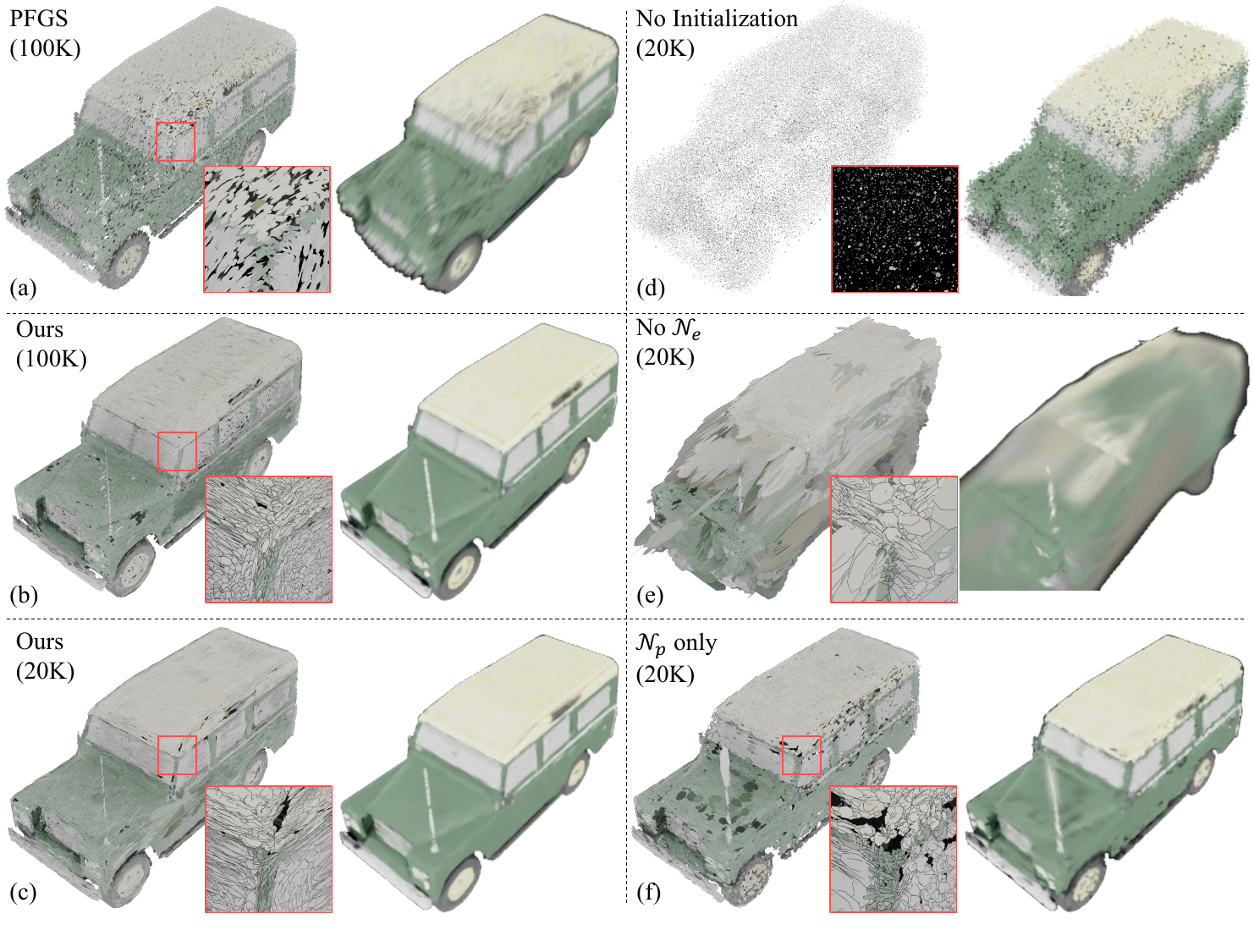}
    \caption{The visualization of Gaussians and their rendered images predicted by our method, our ablated methods and PFGS. (a): PFGS trained on Car (100K). (b) and (c): Our method trained on Car (100K and 20K). (d), (e) and (f): Our ablated methods trained on Car (20K). }
    \label{fig:vis_gs}
\end{figure}

\subsection{Visualization of Gaussians}
Figure \ref{fig:vis_gs}(a)(b) illustrates the predicted Gaussians and corresponding rendered images of our method and PFGS\cite{PFGS}. 
As mentioned in Section \ref{sec:related_work_1}, PFGS first predicts Gaussians and then refines the rendered images. 
We visualize the Gaussians of its first stage.
Different from our predicted Gaussians, which are evenly distributed with fewer gaps, the predicted Gaussians of PFGS are chaos. 
Its rendered image is not photo-realistic requiring refinement in the second stage, and thus cannot be considered a final rendering result.

\subsection{Evaluation of Rendering Speed}
Table \ref{tab:rendering_speed} represents the rendering speeds of different methods.
Our method directly predicts Gaussians for rendering images, eliminating the need for further refinement once the prediction is complete.
The rendering speed of our method is the same as 2DGS \cite{2DGS}, thus our method achieves real-time rendering with the fastest rendering speed among previous methods.

\begin{table}
\centering
\footnotesize
\caption{The rendering speeds of different methods, where the rendering frames per second (FPS) are presented. All methods are evaluated using an RTX 3090.}
\label{tab:rendering_speed}
\begin{tblr}{
  cells = {c},
  hlines,
  vline{2,5} = {-}{},
  hline{1,3} = {-}{0.08em},
}
Method          & NPBG++\cite{NPBG++} & TriVol\cite{TriVol} & PFGS\cite{PFGS} & Ours \\
Speed (FPS) & 37.45  & 1.62   & 3.80 & \textbf{142.86}     
\end{tblr}
\end{table}

\subsection{Ablation Study}
\begin{figure}[t]
    \centering
    \includegraphics[width=\linewidth]{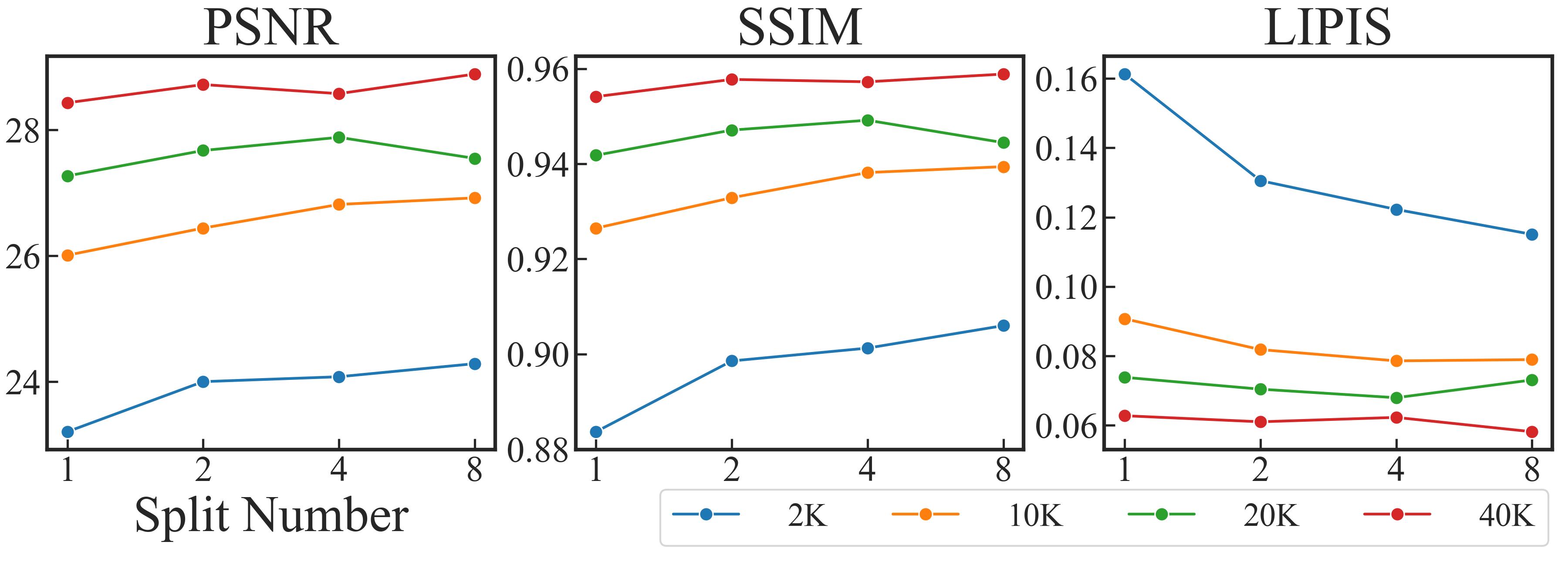}
    \vspace{-1.5em}
    \caption{The ablation study of split number $K$ with different input point numbers. The legend in the lower right corner indicates the input point number for training and evaluating. }
    \vspace{-1.5em}
    \label{fig:ab_split_num}
\end{figure}

\textbf{\textit{Split Number.}}
We conduct experiments with different split numbers $K$ (1, 2, 4, 8) and different input point cloud numbers $N$ (2K, 10K, 20K, 40K) on the Car category of ShapeNet.
The results are shown in Figure \ref{fig:ab_split_num}.
Performance is enhanced with a growing number of splits, underscoring the efficacy of our split decoders, especially when it comes to sparse point clouds.

\noindent\textbf{\textit{Component Ablation.}}
We first evaluate the effectiveness of our initialization in the 2D Gaussian prediction module.
The visualization of initialized Gaussians is shown in Figure \ref{fig:init} (c), which indicates a coarse rendering result providing a great foundation for optimization. 
Without initialization, the performance of our method sharply drops as Table \ref{tab:ablation} shows.
Randomly initialized Gaussians might not be visible in certain views, resulting in non-convergence, as shown in Figure \ref{fig:vis_gs} (d).
The ablation studies of our entire-patch architecture contain two experiments: (1) removing the entire module $\mathcal{N}_e$ and utilizing the predicted incomplete images rendered from point patches for supervision, denoted as ``No $\mathcal{N}_e$'' in Table \ref{tab:ablation}; (2) splitting the entire point cloud to multiple patches as same as the inference process of $\mathcal{N}_p$ to render complete images during training, denoted as ``$\mathcal{N}_p$ only'' in Table \ref{tab:ablation}.
the "$\mathcal{N}_p$ only" method will consume a substantial amount of resources for training and inference.
The evaluation results of these two ablated methods are worse than our method, where their  visualization results are shown in Figure \ref{fig:vis_gs}(e)(f)(c).

The results of ablation studies validate the effectiveness and necessity of the components within our method.

\begin{table}
\centering
\footnotesize
\caption{The ablation studies of our method. The ablated methods are trained on Car with 20K points.}
\vspace{-0.5em}
\label{tab:ablation}
\begin{tblr}{
  cells = {c},0
  column{1} = {r},
  vline{2} = {-}{0.07em},
  hline{3} = {-}{dashed},
  hline{1,6} = {-}{0.12em},
  hline{2,5} = {-}{0.07em},
  colsep = 8pt,
}
                   & PSNR $\uparrow$            & SSIM $\uparrow$            & LPIPS  $\downarrow$          \\
No Initialization      & 9.62           & 0.685          & 0.655          \\
No~$\mathcal{N}_c$ & 12.94          & 0.676          & 0.441          \\
$\mathcal{N}_p$ only     & 23.23          & 0.829          & 0.126          \\
Ours               & \textbf{27.88} & \textbf{0.949} & \textbf{0.068} 
\end{tblr}

\end{table}

\vspace{-0.5em}
\section{Conclusion and Discussion}
In this paper, we introduce a novel approach featuring an entire-patch architecture and the 2D Gaussian prediction module for point cloud rendering.
Our prediction module directly forecasts the 2D Gaussians from the provided point clouds, beginning with initialization derived from normals.
The splitting decoders produce multiple Gaussians from a point, thereby better accommodating sparse point clouds.
To enhance the generalization capability, the entire-patch architecture employs a module for ``background" area prediction and another module for point cloud patch prediction, ensuring proper supervision.
Comprehensive experiments and comparisons demonstrate the superiority and generalization capability of our method, which can render point clouds into photo-realistic images with clear details and achieves SOTA performance.
A limitation of our method may arise when a portion of a point cloud is absent, as our method cannot render a complete image without the direct support of the points.
Our future work aims to address this limitation by incorporating point cloud completion techniques and also attempting to render point clouds with noise, such as those directly scanned from wild and outdoor scenes.

\section*{Acknowledgments}

This work was supported by the National Natural Science Foundation of China under Grant 62032011.

{\small
\bibliographystyle{ieeenat_fullname}
\bibliography{egbib}
}

\end{document}


\title{Supplementary Materials of ``Sparse Point Cloud Patches Rendering via Splitting 2D Gaussians'' }
\author{Changfeng Ma$^1$, Ran Bi$^1$, Jie Guo$^1$, Chongjun Wang$^1$, Yanwen Guo$^{12*}$\\
$^{1}$Nanjing University, Nanjing, China $^{2}$School of Software, North University of China\\
{\tt\small \{changfengma, 211250233\}@smail.nju.edu.cn}\\
{\tt\small \{guojie,chjwang,ywguo\}@nju.edu.cn}
}
\maketitle

\section{Implement Details}

\begin{figure}[t]
    \centering
    \includegraphics[width=\linewidth]{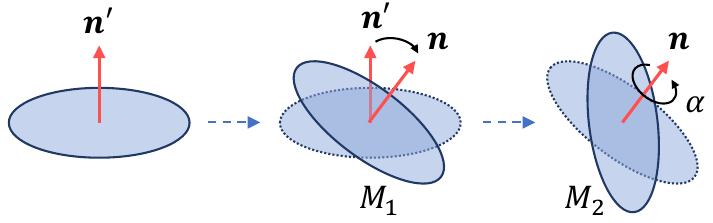}
    \caption{
    The illustration of our conversion process from the normal vector $\mathbf{n}$ and angle $\alpha$ to the quaternion $\mathbf{q}$.
    }
    \label{fig:normal2q}
\end{figure}

\subsection{Quaternion Representation}
In this paper, we utilize the normal vector $\mathbf{n}$ and the rotation angle $\alpha$ about this normal to describe the orientation of Gaussians, rather than employing quaternions as used by 3DGS\cite{3DGS} and 2DGS\cite{2DGS}.
This representation approach simplifies the initialization and prediction of Gaussian rotations in our method.
To facilitate the splatting rendering proposed by 2DGS, we subsequently convert the normal vector $\mathbf{n}$ and the rotation angle $\alpha$ into the quaternion $\mathbf{q}$.
As depicted in Figure \ref{fig:normal2q}, the conversion process consists of two steps.
Firstly, we determine the rotation matrix  $M_1$ from the initial normal $\mathbf{n}'=[0, 0, 1]$ to the predicted normal $\mathbf{n}$.
Subsequently, we compute the rotation matrix $M_2$ that represents the rotation around the normal vector $\mathbf{n}$ by an angle $\alpha$.
These two processes can be accomplished using Rodrigues' rotation formula.
Finally, the quaternion $\mathbf{q}$ is converted from the rotation matrix $M=M_1M_2$.

\subsection{Normalization and De-normalization}
The normalization and de-normalization processes are as follows in the pseudocode below.
\begin{lstlisting}
def normalize(points):
    maxp = max(points, axis=0)
    minp = min(points, axis=0)
    c = (maxp + minp) / 2
    s = max(maxp-c)
    points = (points-c)/s
    return points, c, s
def denormalize(X, S, c, s):
    X = X * s + c
    S = S * s
    return X, S
\end{lstlisting}

\subsection{2D Gaussian Prediction Module}
\textbf{\textit{Encoder.}} 
We employ the basic PointMLP\cite{PointMLP} as our encoder $E$ to extract features $F_l$ from points.
The input channel number of $E$ is $11=3+3+3+2$ (positions $X^i$ (3), colors $C^i$ (3), normals $N^i$ (3) and initialized scale $S^i$ (2)).
The output channel number that is the channel of the features is $640$, where each point contains one feature.

\noindent\textbf{\textit{Splitting Decoder.}}
We utilize a weight-shared Multi-Layer Perceptron (MLP) \cite{2017PointNet} to complement our splitting decoder.
Each parameter of the 2D Gaussian is predicted by a splitting decoder. 
The input channel numbers of all the splitting decoders are $640+11$ (feature $F_l$ (640), positions $X^i$ (3), colors $C^i$ (3), normals $N^i$ (3) and initialized scale $S^i$ (2)).
The hidden layers of all the decoders are 512, 512, 512, 256 and 128.
The output channels of the decoder are $K \times c$, where $K$ is the number of splits and $c$ varies depending on the specific decoder.
The values of $c$ for the decoders $D_x, D_s, D_c, D_n, D_\alpha$ and $D_o$ are 3, 2, 27, 3, 1 and 1, respectively.
The shape of the output from the decoder is $[N , K\times c]$.
Following the reshape operation, the shape of the output becomes $[N\times K, c]$, thereby achieving the splitting of the predicted Gaussian.
Here is the pseudocode for the splitting decoder, using $D_x$ as an example.
\begin{lstlisting}
def D_x(F_l, X_i, C_i, N_i, S_i):
    shift_x = MLP(F_l, X_i, C_i, N_i, S_i)    
    # [N, K*3]
    shift_x = shift_x.reshape([N*K, 3])
    # [N*K, 3]
    X_p = X_i.reshape([N, 1, 3]).repeat([1, K, 1]).reshape([N*K, 3])
    # [N*K, 3]
    X_p = X_p + shift_x
    return X_p
\end{lstlisting}

\subsection{Entire-Patch Architecture}
As we mentioned in the paper, we initially train $\mathcal{N}_e$ using complete point cloud $\mathcal{P}_e$ and corresponding complete images.
Then, we freeze $\mathcal{N}_e$ in order to train $\mathcal{N}_p$. 
The training process is illustrated in the pseudocode below.
\begin{lstlisting}
N_e.requires_grad = False
optimizer = Optimizer(N_p, lr)
for step in range(step_N):
    P_e, I_gt = Training_Data[step]
    G_e = N_e(P_e)
    P_p, mask = get_random_patch(P_e)
    G_p = N_p(P_p)
    G = concatenate(G_e[~mask], G_p[mask]) 
    I_pred = splatting_render(G)
    loss = L(I_pred, I_gt)
    optimizer.zero_grad()
    loss.backward()
    optimizer.step()
\end{lstlisting}

The pseudocode for sampling the entire point cloud $\mathcal{P}_e$ into multiple patches, as utilized in the inference of $\mathcal{N}_p$, is provided below.
\begin{lstlisting}
def get_patches(P_e):
    P_rest = P_e
    P_p = []
    while True:
        center = random_select(P_rest)
        mask = KNN(P_e, center, patch_N)
        p = P_e[mask]
        P_p.append(p)
        P_rest = P_rest[~mask]
        if P_rest is empty:
            break
\end{lstlisting}

\section{More Experiments and Results}

\subsection{Evaluation on Different Point Number}
\begin{figure}[t]
    \centering
    \includegraphics[width=\linewidth]{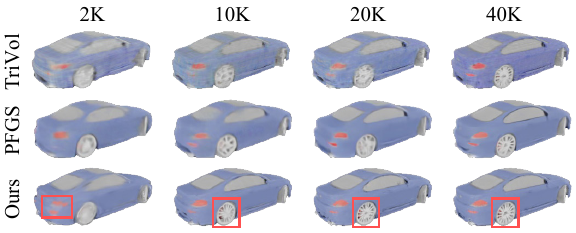}
    \caption{The rendering results of different methods trained on the Car category with 2K, 10K, 20K and 40K points.  }
    \label{fig:vis_dpn}
\end{figure}

Figure \ref{fig:vis_dpn} presents the rendering images of our method, TriVol \cite{TriVol} and PFGS \cite{PFGS} on different point numbers for qualitative comparison.
Here, the methods are trained using Car category with 2K, 10K, 20K and 40K points.
Our rendering results maintain clear details on sparse point clouds.
The taillights predicted by our method exhibit two distinct lights even when the input point cloud contains only 2K points.
Likewise, the wheels predicted by our method are clearer compared to those predicted by other methods.
This comparison illustrates that our method adeptly handles sparse point clouds with the aid of splitting decoders.

\begin{figure}[t]
    \centering
    \includegraphics[width=\linewidth]{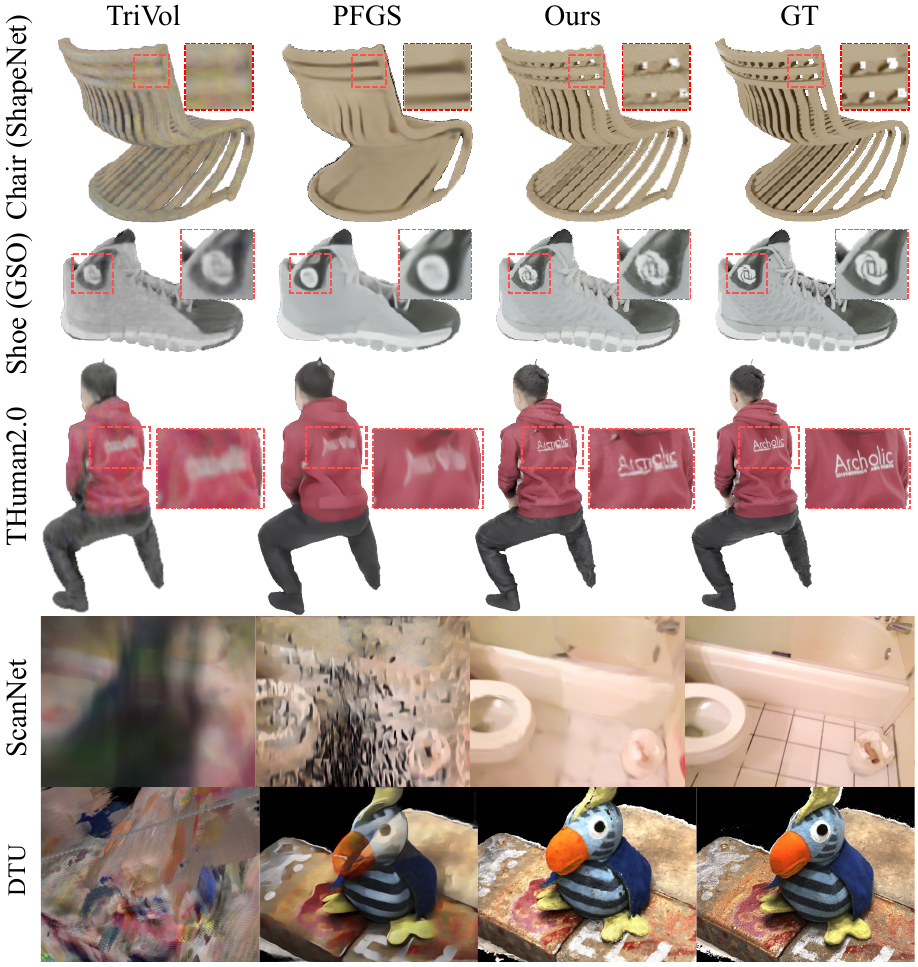}
    \caption{The evaluation results of our methods and previous methods on different datasets, where all methods are trained on the Car category with 20K points.}
    \label{fig:gener_dc}
\end{figure}

\begin{table*}
\footnotesize
\centering
\caption{The evaluation of our method on different datasets including scenes, objects and human bodies, where our method is trained on the Car category with 20K point number.
}
\label{tab:main_com_dc}
\begin{tblr}{
  cells = {c},
  cell{1}{1} = {r=2}{},
  cell{1}{2} = {r=2}{},
  cell{1}{3} = {c=3}{},
  cell{1}{6} = {c=3}{},
  cell{1}{9} = {c=3}{},
  cell{1}{12} = {c=3}{},
  cell{1}{15} = {c=3}{},
  cell{3}{1} = {r=3}{},
  vline{2,3,6,9,12,15} = {-}{0.07em},
  hline{1,6} = {-}{0.12em},
  hline{2} = {3-17}{0.07em},
  hline{3} = {-}{0.07em},
  colsep = 1.75pt,
}
Method & {Point\\Number} & ScanNet\cite{ScanNet} &       &       & Car (ShapeNet\cite{ShapeNet}) &       &       & Chair (ShapeNet\cite{ShapeNet}) &       &       & Shoe (GSO\cite{GSO}) &       &       & THuman2.0\cite{THuman} &       &       \\
       &                 & PSNR$\uparrow$     & SSIM$\uparrow$  & LPIPS$\downarrow$  & PSNR$\uparrow$           & SSIM$\uparrow$  & LPIPS$\downarrow$ & PSNR$\uparrow$             & SSIM$\uparrow$  & LPIPS$\downarrow$ & PSNR$\uparrow$       & SSIM$\uparrow$  & LPIPS$\downarrow$ & PSNR$\uparrow$      & SSIM$\uparrow$  & LPIPS$\downarrow$ \\
{Ours \\ Car(ShapeNet) \\ 20K points} & 20K  & 16.98 & 0.659 & 0.623 & 25.13 & 0.934 & 0.080 & 26.57 & 0.940 & 0.074 & 28.22 & 0.952 & 0.052 & 30.29 & 0.963 & 0.049 \\
     & 40K  & 17.59 & 0.669 & 0.591 & 24.46 & 0.928 & 0.076 & 26.28 & 0.942 & 0.077 & 27.15 & 0.948 & 0.047 & 30.87 & 0.968 & 0.038 \\
     & 100K & 17.86 & 0.672 & 0.560 & 26.52     & 0.931     & 0.050     & 25.25     & 0.935     & 0.060     & 27.18     & 0.949     & 0.037     & 29.58 & 0.966 & 0.033 
\end{tblr}
\end{table*}

\begin{figure}[t]
    \centering
    \includegraphics[width=\linewidth]{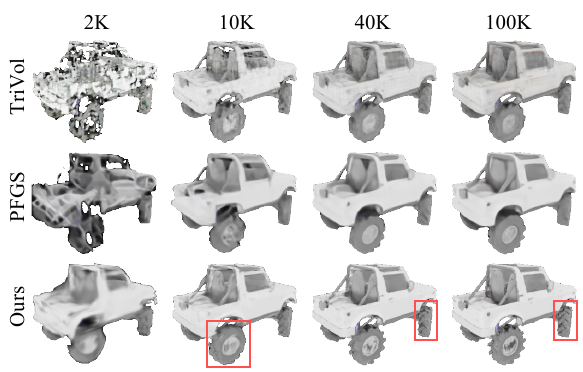}
    \caption{The evaluation results of our methods and previous methods on the Car category with 2K, 10K, 40K and 100K points, where all methods are trained on the Car category with 20K points.}
    \label{fig:gener_dpn}
\end{figure}

\subsection{Evaluation of Generalization Capability}
For a qualitative comparison, we train our method, TriVol, and PFGS on the car category using 20K points, and subsequently evaluate their performance across different categories.
Figure \ref{fig:gener_dc} depicts the rendering outcomes of the methods on additional categories.
These additional categories encompass object-level categories such as chair (ShapeNet \cite{ShapeNet}), shoe (ShapeNet), and human (THuman2.0 \cite{THuman}), as well as scene-level datasets including ScanNet \cite{ScanNet} and DTU \cite{DTU}.
The rendered results on objects of our method exhibit intricate details, such as the lattice of the chair, the pattern of the shoe, and the text on the clothing.
Conversely, other methods only predict blurred results.
Additionally, our method is also applicable to scene-level data, whereas other methods fail to produce accurate results.
Table \ref{tab:main_com_dc} shows evaluation results of our method trained on Car category with 20K on other datasets with varying input point numbers.
Although the performance of our method declined slightly, it was still comparable to the performance of previous works.

As depicted in Figure \ref{fig:gener_dpn}, we also compared the trained methods on the Car category with varying point numbers to verify the robustness of the methods across different point clouds.
Here, the methods are evaluated on 2K, 10K, 40K and 100K points.
The results of our method not only exhibit more details when the point number is high, but also maintain the basic details of objects when the point number is low.
On the contrary, when the point number is high, the predictions of other methods are blurry and lack detail, while they struggle to generate complete images when the point number is low.

Both quantitative and qualitative evaluations demonstrate the exceptional generalization capability of our method across different categories and its robustness on different input point numbers.

\subsection{Multi-View Consistency}
We also render consecutive views around objects to verify the multi-view consistency of different methods.
The results are presented in the video included in our supplementary materials, demonstrating that our method exhibits excellent multi-view consistency and rich detail.
PFGS employs a 2-stage image refinement process to enhance the rendered images of predicted Gaussians, hence its results lack multi-view consistency and exhibit noticeable abrupt changes in the imagery.

\subsection{More Comparison}
\begin{figure*}[t]
    \centering
    \includegraphics[width=\linewidth]{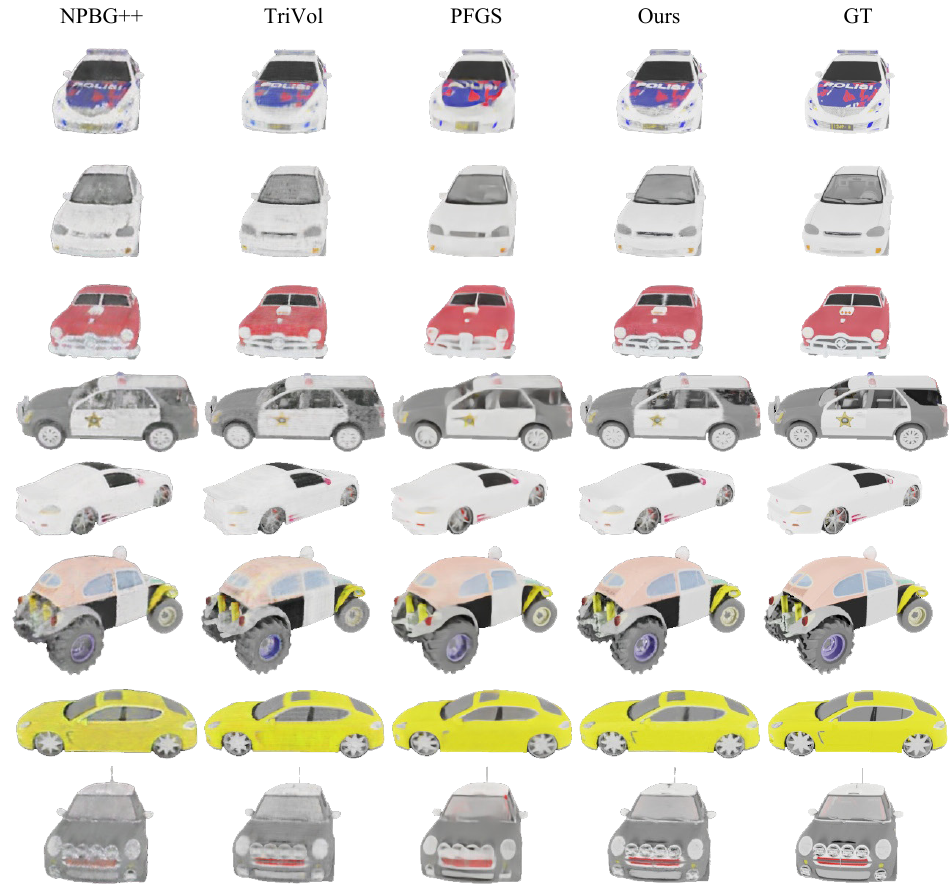}
    \caption{Additional comparison on Car category.}
    \label{fig:com_sup_1}
\end{figure*}

\begin{figure*}[t]
    \centering
    \includegraphics[width=\linewidth]{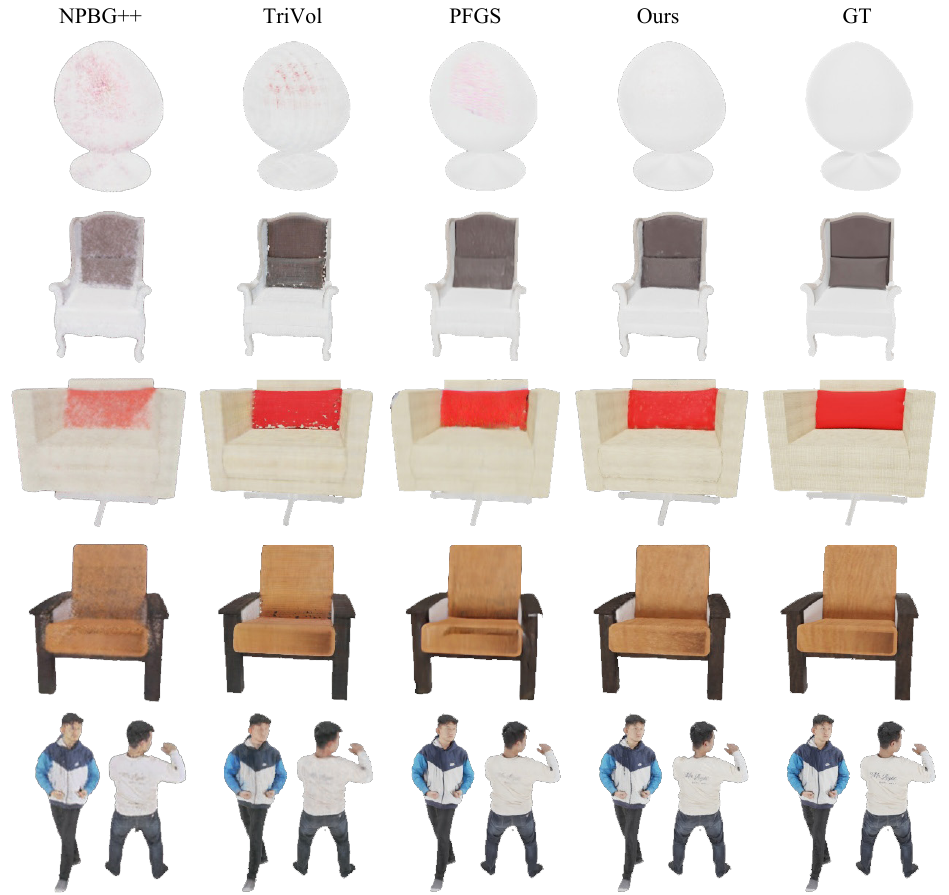}
    \caption{Additional comparison on Chair and Human categories.}
    \label{fig:com_sup_2}
\end{figure*}

\begin{figure*}[t]
    \centering
    \includegraphics[width=\linewidth]{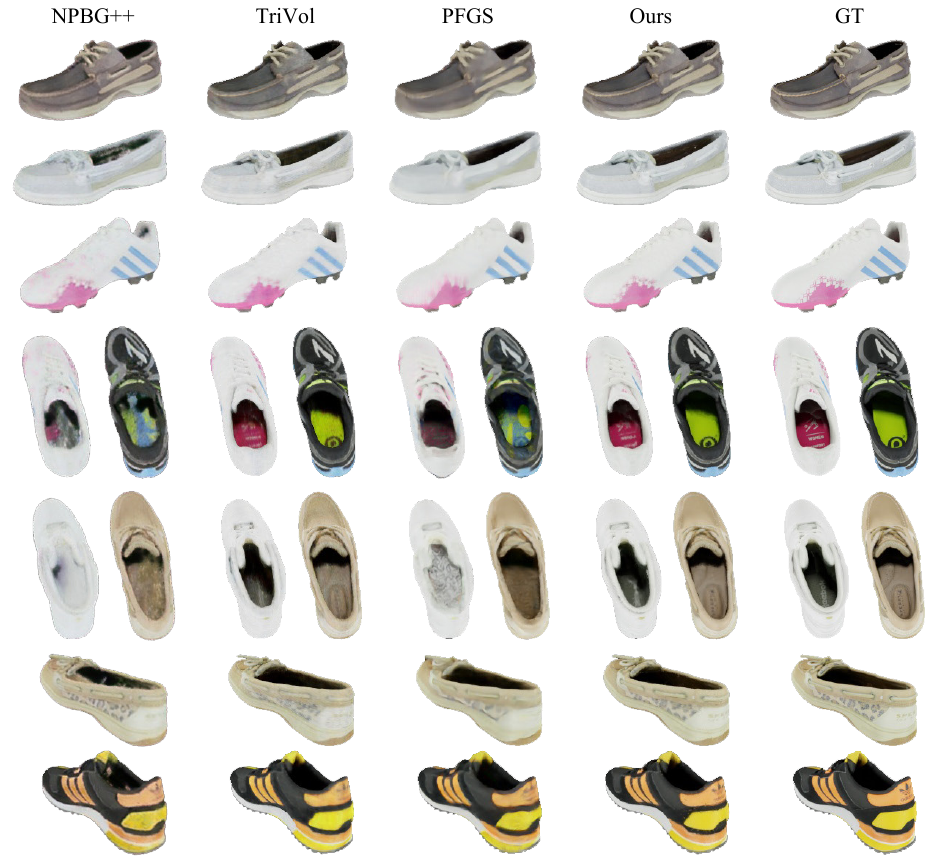}
    \caption{Additional comparison on Shoe category.}
    \label{fig:com_sup_3}
\end{figure*}

\begin{figure*}[t]
    \centering
    \includegraphics[width=\linewidth]{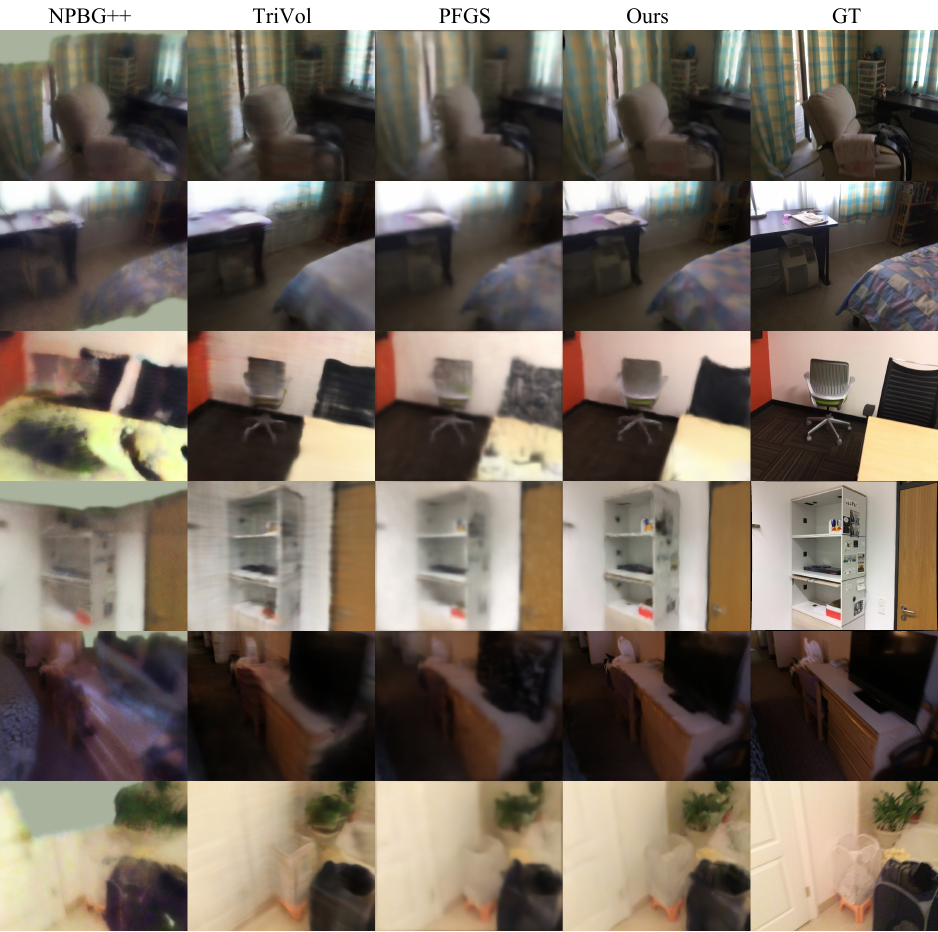}
    \caption{Additional comparison on ScanNet.}
    \label{fig:com_sup_4}
\end{figure*}

Figures \ref{fig:com_sup_1}, \ref{fig:com_sup_2}, \ref{fig:com_sup_3}, and \ref{fig:com_sup_4} present additional comparisons of our method with NPBG++\cite{NPBG++}, TriVol\cite{TriVol}, and PFGS\cite{PFGS}.

\begin{figure}[t]
    \centering
    \includegraphics[width=\linewidth]{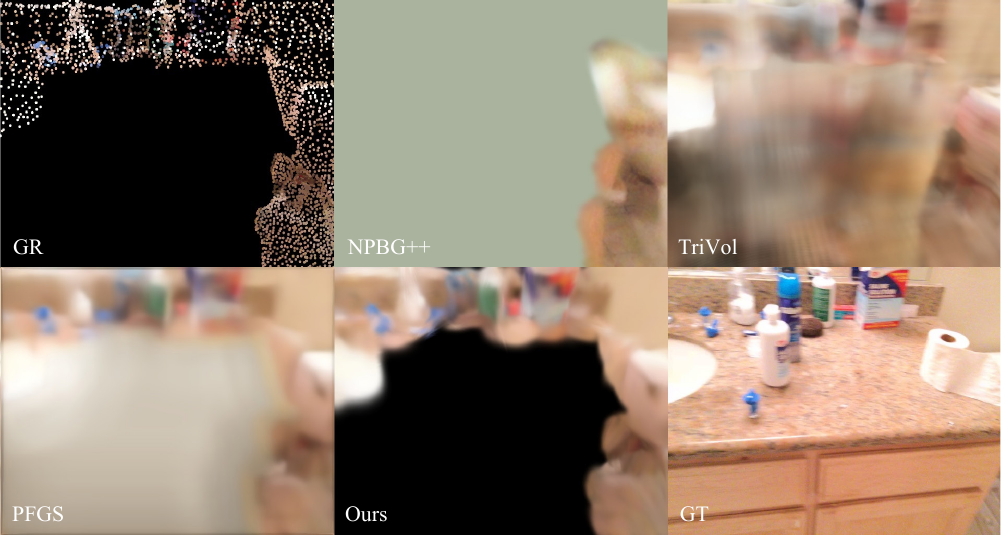}
    \caption{The illustration of our limitation.}
    \label{fig:limitation}
\end{figure}

\subsection{Limitation}
A limitation of our method may arise when a portion of a point cloud is absent.
As depicted in Figure \ref{fig:limitation}, our method is unable to render a complete image without the direct support of the points, leaving areas unfilled and appearing as black.
Such large missing areas are also beyond the predictive capabilities of other methods.
Therefore, our future work aims to address this limitation by incorporating point cloud completion techniques.

{\small
\bibliographystyle{ieeenat_fullname}
\bibliography{egbib}
}